\pdfoutput=1
\documentclass[sigconf]{acmart}

\renewcommand\footnotetextcopyrightpermission[1]{} % removes footnote with conference information in first column

\setcopyright{rightsretained}
\copyrightyear{2024}
\acmYear{2024}
\acmDOI{}

\acmConference[Conference acronym 'XX]{Make sure to enter the correct
  conference title from your rights confirmation emai}{June 03--05,
  2018}{Woodstock, NY}

\acmISBN{}

\begin{document}

\title{How We Refute Claims: Automatic Fact-Checking through Flaw Identification and Explanation}

\author{Wei-Yu Kao}
\affiliation{%
  \institution{National Yang Ming Chiao Tung University}
  \country{Hsinchu, Taiwan}
}
\email{wayner.cs09@nycu.edu.tw}

\author{An-Zi Yen}
\affiliation{%
  \institution{National Yang Ming Chiao Tung University}
  \country{Hsinchu, Taiwan}
}
\email{azyen@nycu.edu.tw}

\begin{abstract}
Automated fact-checking is a crucial task in the governance of internet content. 
Although various studies utilize advanced models to tackle this issue, a significant gap persists in addressing complex real-world rumors and deceptive claims. 
To address this challenge, this paper explores the novel task of flaw-oriented fact-checking, including aspect generation and flaw identification.
We also introduce RefuteClaim, a new framework designed specifically for this task.
Given the absence of an existing dataset, we present FlawCheck, a dataset created by extracting and transforming insights from expert reviews into relevant aspects and identified flaws. %designed to encompass crucial aspects and seven critical flaws identified by fact-checking experts. 
The experimental results underscore the efficacy of RefuteClaim, particularly in classifying and elucidating false claims.
\end{abstract}

%% https://dl.acm.org/ccs#


%Natural Language Explanation, Web Text Classification, Large Language Model}

\maketitle

\section{Introduction}\label{sec:intro}
As the World Wide Web continues to expand, an overwhelming amount of information is flooding the internet. 
With limited time available, individuals often struggle to comprehensively grasp content and may be adversely influenced by problematic claims.
Accurately assessing the veracity of a claim demands considerable time and effort from fact-checking organizations and experts, who meticulously gather relevant evidence and scrutinize the facts and assumptions. 
Hence, in the quest to efficiently identify a plethora of false statements on the internet, numerous studies have investigated automatic fact-checking.

Research on automatic fact-checking can be broadly categorized into veracity classification and justification generation, with a predominant focus on the former. 
Although veracity classification is essential for prompt assessment of claims, justification generation plays a more pivotal role in producing explainable results and facilitating comprehensive fact-checking.
To generate justifications in natural language, early studies~\cite{shu2019defend, atanasova2020generating} proposed directly extracting sentences from reliable sources such as news content.
More recent work~\cite{chen2022generating, rani2023factify, pan2023fact} explores the approach of question answering,
which involves generating questions related to specific facts within the claims. 
By answering these generated questions using supporting evidence, models predict and articulate the authenticity of the claims, fostering more robust fact-checking.

Despite substantial advancements in the field, scant attention has been directed toward discerning the underlying causes that render a claim false. 
For instance, in the question answering approach, the generated questions predominantly focus on straightforward facts such as who, when, where, and how.
In real-world scenarios, deceptive claims are often intricately crafted to mislead without falling into obvious pitfalls, strategically distorting a minor segment of the claim. 
This is precisely why professional fact-checkers thoughtfully analyze claims from diverse perspectives, assessing not only the literal accuracy of the statement but also considering elements such as tone, context, and its consistency with established facts.
Thus, it becomes imperative to identify the specific flaws within a claim to comprehensively confirm and elucidate its veracity in an automatic fact-checking process.
 
To address the challenge at hand, we first employ aspect generation to determine the most crucial aspects associated with the claim, around which evaluation should be focused.
Based on evidence, our model synthesizes explanations for these distinct aspects.
Then, we carefully select seven critical flaws from an initial set of twenty-two false labels used by various fact-checking organizations to debunk claims. 
These flaws are grouped into three distinct categories, each influencing the outcome of the fact-checking process in unique ways.
The first category includes three explicit flaws that can be identified by evaluating the claim's compatibility with the evidence: \textbf{contradicting facts}, \textbf{exaggeration}, and \textbf{understatement}.
In the second category, we introduce two more nuanced flaws: \textbf{occasional faltering} and \textbf{insufficient support}.
Fact-checkers working within this realm must employ critical reasoning to envisage potential scenarios where the claim may not hold true.
A claim may seem convincing initially but fail to maintain its validity or soundness over time and across different scenarios.
The final category encapsulates the two most complex flaws: \textbf{problematic assumptions} and \textbf{existence of alternative explanations}. 
Identifying these particular flaws is a more intricate process, as it requires fact-checkers to consider a wider context and often demands extensive background knowledge that may not be demonstrated in the evidence at hand.
Details will be elaborated in the following section.

In the absence of an existing dataset, we extend WatClaimCheck~\cite{khan2022watclaimcheck}, utilizing a large language model (LLM) to infer the aspects and flaws of claims based on review articles, thereby constructing the FlawCheck dataset.\footnote{\url{https://github.com/NYCU-NLP-Lab/FlawCheck.git}}
Subsequently, we present RefuteClaim, a novel framework designed to guide the review process in creating comprehensive fact-checking articles to emulate the quality and depth of articles generated by human experts. 

In summary, our contributions are summarized as follows:

\begin{itemize} 
    \item We introduce a novel flaw-checking task that entails the examination of the seven flaws, reflecting the complexities of real-world automatic fact-checking scenarios.
    \item We present the FlawCheck dataset, which encompasses distinct aspects and explanations for the seven flaws associated with each claim.
    This dataset encapsulates the expertise of human fact-checking professionals.
    \item We propose RefuteClaim, a framework that integrates aspect generation and flaw identification into an automatic fact-checking pipeline.
    Experimental results show promising performance, both in classifying and elucidating false claims.
\end{itemize}

\section{Flaw-Oriented Fact-Checking}
In this study, we develop RefuteClaim, which incorporates aspect generation along with flaw identification and explanation for fact-checking.
The definitions are provided as follows. 

\subsection{Aspect Generation}
Aspect generation aids in identifying and focusing on the key elements around which the evaluation of a claim revolves. 
The goal is to determine the most crucial aspects associated with a claim to guide subsequent flaw identification and explanation. 
Aspects represent the specific dimensions or attributes that are integral to evaluating the validity of a claim,
for example, a statement asserting that a particular politician engaged in corrupt practices during an election campaign. 
In this context, aspects could include:

\noindent \textbf{Legal investigations:} Evaluating whether there are ongoing or concluded legal investigations into the alleged corrupt practices.

\noindent \textbf{Financial transactions:} Scrutinizing financial transactions related to the campaign to identify any irregular activities.

\noindent \textbf{Political motivations:} Investigating if there are political motivations behind the accusations, such as rivalry strategies.

\noindent \textbf{Legal precedents:} Examining similar cases in the past to understand how they were adjudicated.

The relevant evidence associated with a claim may lead to various aspects.
In this work, our model explains up to four distinct aspects.

\subsection{Flaw Identification and Explanation}
The process of flaw identification and explanation involves examining statements or claims critically to identify inaccuracies or logical inconsistencies. 
Here we define and detail seven types of flaws that may be present in a claim:

\noindent \textbf{Contradicting facts:}
When a claim is presented, it must align with established facts. 
A contradicting fact occurs when the claim directly opposes known and verified information. 
For instance, the claim that ``the Earth is flat'' contradicts the overwhelming scientific evidence that the Earth is an oblate spheroid. 
It is crucial to cross-reference claims with reliable data sources to detect such flaws.

\noindent \textbf{Exaggeration:}
Exaggeration is a flaw where the truth is stretched beyond its actual proportions. 
This can be done to make something appear more significant or severe than it really is. 
For example, stating that ``everyone hates the new policy'' is likely an exaggeration, as it is improbable that every single person has a negative view.

\noindent \textbf{Understatement:}
An understatement minimizes the significance of something in a way that misrepresents the truth and downplays important facts. 
Saying ``climate change is a minor issue'' is an understatement because it fails to convey the widespread consensus on the seriousness of climate change impacts.

\noindent \textbf{Occasional faltering:}
A claim falters at times when it is presented as universally true, but cannot be consistently sustained.  
For example, the claim that ``electric cars are always better for the environment'' may falter in regions where electricity is primarily generated from coal, potentially making the environmental benefits less clear-cut. 
This inconsistency reveals that the claim does not account for specific conditions where it may not hold true.

\noindent \textbf{Insufficient support:}
Claims should be backed up by evidence. 
Support is insufficient when assertions are made without the necessary substantiation. 
For instance, claiming that ``a particular diet causes weight loss in all individuals'' without citing scientific studies or statistical data is an unsupported statement.

\noindent \textbf{Problematic assumptions:}
This type of flaw arises when a claim is based on assumptions that are not validated or are questionable.
Problematic assumptions can lead to incorrect conclusions.
An example would be assuming that ``increased internet usage directly causes poor social skills,'' which ignores potential factors like the type of internet usage or individual differences.

\noindent \textbf{Existence of alternative explanations:}
Even if a claim seems plausible, other explanations may account for the observed facts. 
A flaw exists when a claim does not consider or rule out such alternative explanations. 
For example, the conclusion that ``rising smartphone sales are due solely to improved technology'' ignores other possible factors such as marketing strategies or changes in consumer behavior.

Note that a single claim may incorporate several distinct flaws, and the composition of aspects is indeed diverse.
\label{sec: flaws}

\section{Dataset Construction}
Due to the demand for premise articles and complete review articles written by human experts, we selected WatClaimCheck~\cite{khan2022watclaimcheck} as our data source, given its ample and varied collection from eight fact-checking websites. 
In this study, we extend 33,721 claims and their metadata in WatClaimCheck to facilitate the proposed flaw-oriented fact-checking task, leading to the creation of FlawCheck.

Since the original content in WatClaimCheck includes a significant amount of irrelevant web crawl data, we collected the web data again and cleaned it to ensure relatively clean review articles for the evaluation of justification generation.
Then, we harness the capabilities of \texttt{GPT-3.5-turbo} to distill expert opinions from review articles and transform them into the various aspects and identify flaws. 
Given an input claim $C_i$ and a review article $R_i$, we initially generate four silver ground-truth aspects $\mathcal{A}_i=\{A_{i,1}, A_{i,2}, \ldots, A_{i,n}\}$, where $1 \leq n \leq 4$. 
The aspects may represent coarse-grained facts referenced by the claim. 
Following the same process, taking $C_i$ and $R_i$, we utilize \texttt{GPT-3.5-turbo} to transform the human expert argumentation to elaborate the presence of flaws $\mathcal{F}_i=\{F_{i,1}, F_{i,2}, \ldots, F_{i,7}\}$.

Furthermore, to facilitate the evaluation of veracity classification, we reassigned the labels in WatClaimCheck due to inaccuracies in their arrangement, for instance, classifying the label ``Pants on Fire'' to ``Partially True/False.''
We define four labels---Incorrect, Partly false, Unproven, and Correct---based on the original ratings obtained from each fact-checking website.
FlawCheck's label distribution in the training and testing sets is as follows: ``True'' with 5,272 and 657, ``Unproven'' with 805 and 112, ``Partly false'' with 3,429 and 451, and ``False'' with 17,470 and 2,153 instances, respectively. 
This imbalanced distribution of labels is due primarily to the nature of fact-checking websites.

\section{Methodology}
In this section, we present the four integral components of the proposed RefuteClaim framework: 
evidence retriever, aspect generator, flaw checker, and justification generator. 

\paragraph{Evidence retriever.} 
We utilize the widely-adopted dense passage retriever (DPR)~\citep{karpukhin-etal-2020-dense}. 
Additionally, per \citet{khan2022watclaimcheck}, we employ review articles as the source of truth when training the DPR model.
We segment the review articles into sentences, represented as $R_i = \{r_{i,1}, r_{i,2}, \ldots, r_{i,n}\}$, where $n$ is the number of sentences in the $i$-th review article.
Let $R^{+}_i$ = \{$\mathit{rs}^{+}_{i,1}, \ldots, \mathit{rs}^{+}_{i,\alpha}$\}, $R^{-}_i$ = \{$\mathit{rs}^{-}_{j,1}, \ldots, \mathit{rs}^{-}_{j,\beta}$\} denote the set of positive and negative examples of the $i$-th claim $C_i$, respectively.
The training data is $\{\langle C_i, R^{+}_i, R^{-}_i \rangle \}^{d}_{i=1}$, which contains $d$ instances.
Each instance consists of one claim $C_i$, $\alpha$ positive examples $\mathit{rs}^{+}_{i,q}$, and $\beta$ negative examples $\mathit{rs}^{-}_{j,r}$.
During the training phase, we minimize the negative log likelihood loss function as
\begin{equation}
    %\footnotesize
     L = -\log \frac{e^{\mathrm{sim}(C_i, \mathit{rs}^{+}_{i,q})}}{\sum_{q=1}^{\alpha}e^{\mathrm{sim}(C_i, \mathit{rs}^{+}_{i,q})}+\sum_{r=1}^{\beta}e^{\mathrm{sim}(C_i, \mathit{rs}^{-}_{j,r})}},
\end{equation}
where $\mathrm{sim}(v, u)$ is the dot product between claim vector $v$ and sentence vector $u$.
Using the evidence retriever, we extract 50 sentences from the claim's premise articles to serve as evidence $E_i = \{e_{i,1}, e_{i,2}, \ldots, e_{i,50}\}$.

\paragraph{Aspect generator.}
We employ the aspect generator to generate up to four key aspects, encapsulating the most critical and diverse directions to fact-check the claim.
For each sequence-to-sequence model, we employ LoRA fine-tuning~\cite{hu2021lora} to train an LLM $\mathcal{M}$, parameterized by $\Phi$.
The training data used in LoRA fine-tuning can be denoted as $\mathcal{Z} = \{(x_i, y_i)\}_{i=1, \ldots, N}$. 
Given task-specific parameter increment $\Delta\Phi = \Delta\Phi(\Theta)$, we optimize over $\Theta$:
\begin{equation}
    \label{eq:lora}
    %\footnotesize
    \max_{\Theta} \sum_{(x,y) \in \mathcal{Z}} \sum_{t=1}^{|y|} (\log p_{{\Phi}_0 + \Delta\Phi(\Theta)}(y_t | x, y_{<t})).
\end{equation}
Where $\mathcal{P}_A$ is the prompt that takes $C_i$ and $E_i$ to generate aspects, we have $x_i = \mathcal{P}_A(C_i, E_i)$ and $y_i = \mathcal{A}_i$.

\paragraph{Flaw checker.}
The flaw checker identifies and explains the presence or absence of the flaws from Section~\ref{sec: flaws}.
The optimization and utilization of the model adhere to Equation~\ref{eq:lora}, where $\mathcal{P}_F$ is the prompt that takes $C_i$, $\mathcal{A}_i$, and $E_i$ to flaw-check.
Finally, $x_i = \mathcal{P}_F(C_i, \mathcal{A}_i, E_i)$ and $y_i = \mathcal{F}_i$.

\paragraph{Justification generator.}
The justification generator assembles $C_i$, $\mathcal{F}_i$, and $E_i$ to generate the final review for the claims. 
As the flaw checker already incorporates aspects in generating its output, to optimize model input space, we here omit the aspect as an additional input.

\section{Experiments}
\begin{table*}
  \scriptsize
  \setlength\tabcolsep{3.5pt}
  \begin{tabular}{l|cccc|cccc|cccc|cccc}\toprule
    &\multicolumn{4}{c|}{ROUGE-1} &\multicolumn{4}{c|}{ROUGE-2} &\multicolumn{4}{c|}{ROUGE-L} &\multicolumn{4}{c}{BERTScore} \\\cmidrule{2-17}
    Models &False &Partly false &Unproven &True &False &Partly false &Unproven &True &False &Partly false &Unproven &True &False &Partly false &Unproven &True \\
    \midrule
    Baseline &0.3151 &0.2709 &\textbf{0.3107} &0.3091 &0.1087 &0.0887 &\textbf{0.1089} &\textbf{0.1117} &0.1644 &0.1355 &\textbf{0.1697} &0.1629 &0.8236 &0.8212 &\textbf{0.8266} &0.8269 \\
    \hspace{1.5mm} w/ aspects &0.3176 &0.2721 &0.2935 &0.3020 &\textbf{0.1128} &0.0932 &0.0993 &0.1116 &0.1711 &0.1392 &0.1608 &0.1623 &\textbf{0.8268} &\textbf{0.8246} &0.8264 &\textbf{0.8274} \\
    \midrule
    RefuteClaim-3F &0.3119 &0.2713 &0.2836 &0.3010 &0.1033 &0.0828 &0.0873 &0.1060 &0.1683 &0.1397 &0.1577 &0.1643 &0.8215 &0.8175 &0.8167 &0.8217 \\
    RefuteClaim-5F &0.3235 &\textbf{0.2828} &0.2994 &0.3055 &0.1122 &\textbf{0.0948} &0.1004 &0.1046 &0.1726 &0.1432 &0.1641 &0.1605 &0.8261 &0.8214 &0.8224 &0.8243 \\
    RefuteClaim-7F &\textbf{0.3266} &0.2788 &0.2838 &\textbf{0.3106} &0.1109 &0.0868 &0.0832 &0.1091 &\textbf{0.1739} &\textbf{0.1433} &0.1493 &\textbf{0.1682} &0.8245 &0.8183 &0.8179 &0.8243 \\
\bottomrule
\end{tabular}
  \caption{Justification generation}
  \label{tab:justification result}
  %\vspace{-3.0em}
\end{table*}

\begin{table}
  \scriptsize
  \setlength\tabcolsep{1.5pt}
  \begin{tabular}{l|cccc|cccc}\toprule
    &\multicolumn{4}{c|}{Correctness} &\multicolumn{4}{c}{Completeness} \\\cmidrule{2-9}
    Models &False &Partly false &Unproven &True &False &Partly false &Unproven &True \\
    \midrule
    Baseline &0.4770 &0.4700 &0.4311 &0.4675 &0.5165 &0.4790 &0.4644 &0.5031 \\
    \hspace{1.5mm} w/ aspects &0.4400 &0.4780 &\textbf{0.4580} &0.5069 &0.4825 &0.4827 &\textbf{0.4750} &0.5394 \\
    \midrule
    RefuteClaim-3F &0.4540 &0.4288 &0.3778 &0.4825 &0.4870 &0.4475 &0.3889 &0.5300 \\
    RefuteClaim-5F &0.4970 &0.4970 &0.4000 &0.5075 &0.5170 &0.5140 &0.4000 &0.5275 \\
    RefuteClaim-7F &\textbf{0.5088} &\textbf{0.5140} &0.4356 &\textbf{0.5280} &\textbf{0.5381} &\textbf{0.5186} &0.4611 &\textbf{0.5450} \\
    \bottomrule
\end{tabular}
  \caption{Justification evaluation using Gemini Pro}
  %Note that the score doesn't necessarily represent the accuracy but serves as a measure of relative performance. }
  \label{tab:gemini result}
  %\vspace{-3.0em}
\end{table}
% AIzaSyAsk4JHlpM_z1XxiA-ZnAwvWDo4E0TV7Z8

\subsection{Baseline Models}
\textbf{Simple Baseline:}
The simple baseline represents the basic utilization of LLMs, a setting commonly adopted by the majority of related work, which solely considers $C_i$ and $E_i$. 
The model 
% operates independently of assistance from
does not consider aspects or flaw explanations. 

\noindent \textbf{With Aspects:}
Another baseline model takes into account \(\mathcal{A}_i\) to follow specified directions during justification generation.
In the absence of flaw explanations, we evaluate the efficacy of aspect generation. 
This baseline model utilizes $C_i$, $\mathcal{A}_i$, and $E_i$ to directly generate justifications. 

\subsection{Experimental Setup}
In our experiments we utilized \texttt{Vicuna-7b-v1.5}~\citep{zheng2023judging} as the LLM.
During LoRA fine-tuning, we set the rank to 8.
To evaluate the justification results, we adopted ROUGE-1, ROUGE-2, ROUGE-L, and BERTScore as evaluation metrics.
Given the absence of an existing metric to assess critical qualities such as correctness and completeness in fact-checking, we leveraged Google \texttt{Gemini Pro}\footnote{We chose \texttt{Gemini Pro} for its free access, enabling thorough result evaluation.} to score the generated justifications against the ground truth.
\texttt{Gemini Pro} assigns scores on a scale of 0 to~1, where a score of 0 indicates a lack of quality (e.g., correctness), and a score of 1 denotes its full embodiment.

To assess whether the generated justifications contribute to fact-checking, we also evaluated the results of veracity classification.
\citet{hu2023bad} demonstrate the potential in utilizing LLMs to generate fact-checking descriptions.
Nevertheless, such autoregressive models may demonstrate sub-optimal performance in veracity classification.
In addition, to produce strictly deterministic output, we trained a RoBERTa-large model \cite{conneau2020unsupervised} $\mathcal{M}_V$ as our veracity classifier.
We utilized the review articles to train $\mathcal{M}_V$ and employed input justifications generated by different models for veracity classification. 
Developing a proper classifier to connect with different justification generators is left as future work.

\subsection{Experimental Results}
\begin{table}
  \footnotesize
  \begin{tabular}{lccccc}
  \toprule
    &\multicolumn{4}{c}{Accuracy} & \\
    \cmidrule{2-5}
    Data source &False &Partly false &Unproven &True &Macro F1 \\
    \midrule
    Golden review &0.7902 &0.2111 &0.3182 &0.7431 &0.4993 \\
    \midrule
    Baseline &\textbf{0.7439} &0.0889 &0.0682 &0.6166 &0.3621 \\
    \hspace{1.5mm} w/ aspects &0.6780 &0.0722 &0.0682 &\textbf{0.6957} &0.3492 \\
    \midrule
    RefuteClaim-3F &0.7122 &0.0778 &0.0227 &0.5771 &0.3255 \\
    RefuteClaim-5F &0.7183 &0.1278 &\textbf{0.1136} &0.6087 &\textbf{0.3763} \\
    RefuteClaim-7F &\textbf{0.7439} &\textbf{0.1611} &0.0455 &0.5850 &0.3733 \\
\bottomrule
\end{tabular}
  \caption{Veracity classification}
  \label{tab:veracity result}
  %\vspace{-3.0em}
\end{table}

\subsubsection{Justification Generation}
Table \ref{tab:justification result} presents the results using ROUGE scores and BERTScore as metrics.
As mentioned in Section~\ref{sec:intro}, the seven flaws are grouped into three categories.
The RefuteClaim variants 3F, 5F, and 7F denote the first category (three flaws), the first two categories (five flaws), and all three categories (seven flaws), respectively.
The results show that the RefuteClaim models achieve promising performance in ROUGE-1 and ROUGE-L for the ``False'', ``Partly false'', and ``True'' claims, indicating its effectiveness in generating justifications across different veracity levels.
However, our models struggle with generating justifications for ``Unproven'' claims.
Meanwhile, RefuteClaim models under different settings exhibit weaker BERTScore performance, 
perhaps because we defined seven flaw types, and some flaw elucidations generated by \texttt{GPT-3.5-turbo} require inference. 
Consequently, justifications generated based on the output from the flaw checker trained by silver labels could significantly diverge from the ground truth, i.e., review articles. 
Nevertheless, comparison of the baseline models shows that integrating the aspect improves justification results in most cases due to the model's increased focus on several crucial points. 

We evaluate the generated justification quality in Table~\ref{tab:gemini result}. 
RefuteClaim with seven flaws (RefuteClaim-7F) outperforms other methods, except for unproven claims.
The RefuteClaim results with three, five, and seven flaws suggest that a holistic consideration of all flaws, particularly the inclusion of ``problematic assumptions'' and ``existence of alternative explanations'', benefits justification generation.
For unproven claims, the model must focus on more trivial details or find insufficiency in current evidence, whereas our defined flaws focus more on detecting significant errors in claims and introducing more inferences made by \texttt{GPT-3.5-turbo} based on reviews.
Incorporating only aspects is a more fitting approach for unproven claims, as it faithfully represents most content written by experts.

\subsubsection{Veracity Classification}
Table \ref{tab:veracity result} presents the results of veracity classification.
``Golden review'' denotes direct utilization of expert-written review articles to train the veracity classifier: this is challenging.
As shown in Table~\ref{tab:veracity result}, RefuteClaim-7F exhibits the highest macro F1 score, particularly excelling in rating  ``False'' and ``Partly false'' claims compared to other methods. 
Nevertheless, RefuteClaim-5F achieves superior performance in identifying unproven claims.
This aligns intuitively with its incorporation of relevant flaws, i.e., occasional faltering and insufficient support.
Moreover, the RefuteClaim models exhibit diminished accuracy scores when evaluating claims as ``True'' in comparison to the baseline models.
This can be attributed to the nature of RefuteClaim models, which focus on highlighting flaws.
Such an emphasis increases the likelihood of employing negative statements, potentially confusing the classifier.
Furthermore, our classifier exhibits notably poor performance in rating claims as ``Partly false'' and ``Unproven''. 
Even when utilizing the golden review, accuracy does not improve. 
Upon closer examination of misclassified samples, we find that the classifier struggles to discern between ``Partly false'' and ``False'', as well as between ``Unproven'' and ``Partly false''.
Mere incorporation of the aspect or three-flaw modules is not enough to address this challenge. 
However, we believe the flaw identification and generation findings are valuable for the community.

\section{Conclusion and Future Work}
This paper explores the process of forming a flaw-checking perspective to generate justifications in an effort to emulate the quality of fact-checking conducted by human professionals.
We introduce the novel task of flaw-oriented fact-checking and present FlawCheck, a dataset encompassing the critical aspects identified by experts and seven pivotal flaws that demand evaluation.
A pilot framework, RefuteClaim, is proposed.
The experimental results highlight the effectiveness of RefuteClaim in elucidating and classifying false claims. 
As a pioneering work in studying flaw-oriented fact-checking, further investigation is required to explore the optimal utilization of aspects and flaws.

\bibliographystyle{ACM-Reference-Format}
\bibliography{sample-base}

% \appendix

% \section{Appendix}

\end{document}